\documentclass[11pt]{article}
\pdfoutput=1
\usepackage[utf8]{inputenc}
\usepackage[T1]{fontenc}
\usepackage[margin=1in]{geometry}
\usepackage{amsmath,amssymb}
\usepackage{newtxtext,newtxmath}   
\usepackage{booktabs}
\usepackage{graphicx}
\usepackage{float}                 
\usepackage{microtype}
\usepackage{xcolor}
\definecolor{whisslegreen}{HTML}{124E3F}
\definecolor{whisslered}{HTML}{C0392B}
\usepackage{titlesec}
\usepackage{titling}
\usepackage{caption}
\usepackage[colorlinks=true,linkcolor=whisslegreen,citecolor=whisslered,urlcolor=whisslegreen]{hyperref}
\titleformat{\section}{\large\bfseries\color{whisslegreen}}{\thesection}{0.6em}{}
\titleformat{\subsection}{\normalsize\bfseries\color{whisslegreen}}{\thesubsection}{0.6em}{}
\titlespacing*{\section}{0pt}{1.4ex}{0.6ex}
\pretitle{\begin{center}\LARGE\bfseries\color{whisslegreen}}
\posttitle{\par\end{center}\vspace{0.3em}\begin{center}{\color{whisslered}\rule{\textwidth}{1.2pt}}\end{center}\vspace{0.2em}}
\preauthor{\begin{center}\large}
\postauthor{\end{center}}
\predate{\begin{center}\small\color{gray}}
\postdate{\par\end{center}}
\usepackage{fancyhdr}
\begin{document}
\title{Catching Lies Without Sending the Video: Privacy-Preserving Multimodal Deception Detection}
\author{Nikita Sharma\textsuperscript{1} \quad Pranav Saran\textsuperscript{2} \quad Karan Singla\textsuperscript{3}\\[0.5em]
{\normalsize \textsuperscript{1}University of California, San Diego \quad \textsuperscript{2}Case Western Reserve University, Ohio \quad \textsuperscript{3}WhissleAI, USA}}
\date{\today}
\maketitle
\begin{abstract}
Frontier multimodal models can guess whether a person is lying from a testimony video. To do so, they stream that raw face and voice to a third-party model. We ask whether the heavy media is needed at all. On the \emph{Real-life Trial Deception} dataset, Whissle's on-device speech and vision stack extracts a compact digest: transcript, emotion, age, gender, intent distributions, a deception-intent filter, fluency and rhythm, per-frame facial behaviour, and prosody. Under speaker-independent evaluation, we report three findings. A small classifier on this digest reaches \textbf{AUC 0.741}, matching Gemini 2.5 Pro on full video. Handing the digest to a frontier LLM reaches \textbf{AUC 0.755 with Claude Opus 4.8} at \textbf{7.8$\times$ fewer input tokens}, with no media leaving the device. The reported 75\% accuracy is a speaker-leakage artefact. We release code and experiments.\footnote{\url{https://github.com/WhissleAI/lie_detection_binary}}
\end{abstract}

\section{Introduction}

When a witness takes the stand, every pause, glance, and word is scrutinised for deception. Automating that judgement is an old and fraught dream. The polygraph is unreliable and easily countered. Human observers barely beat chance \cite{ref1}. Multimodal large language models (LLMs) have revived the dream. Give a model the video, ask "is this person lying?", and it does meaningfully better than chance \cite{ref6}.

\begin{figure}[H]
\centering
\includegraphics[width=\linewidth]{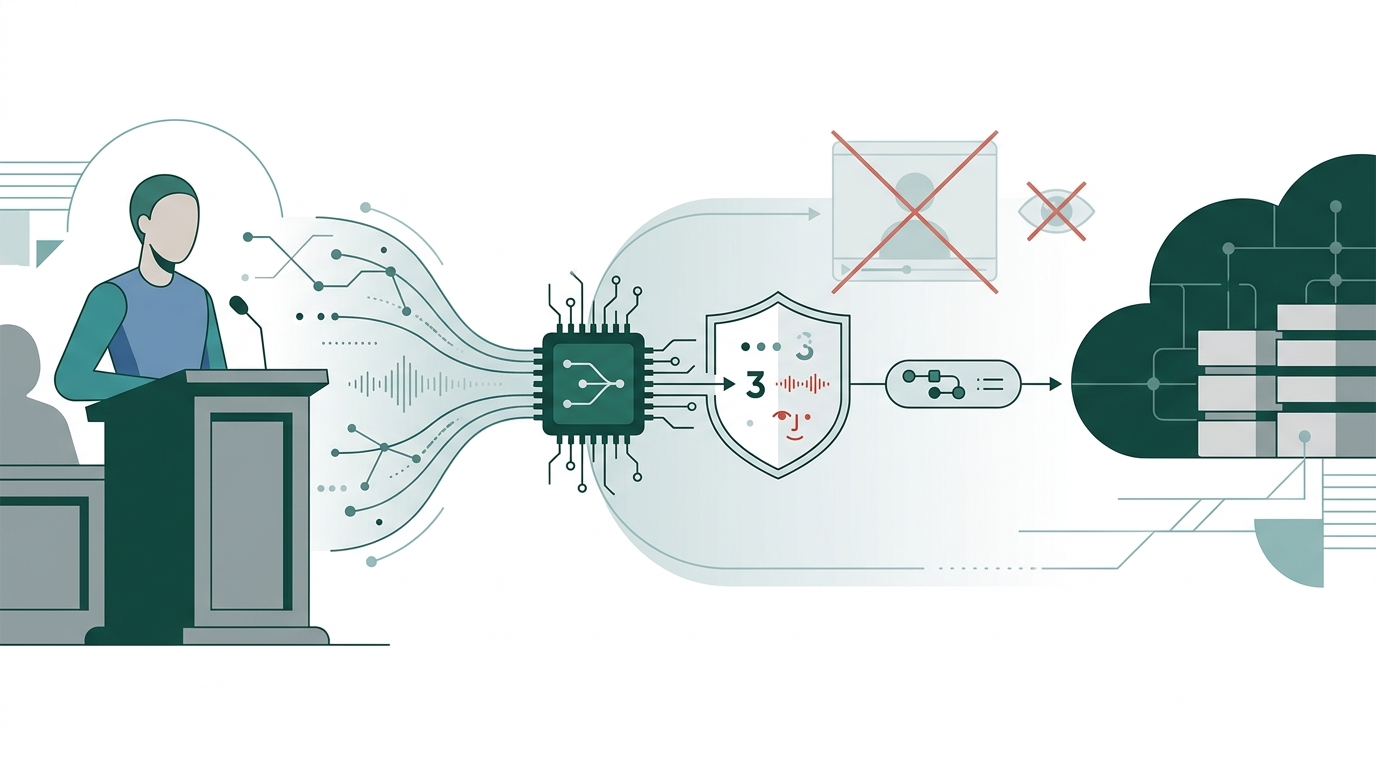}
\caption{On-device, a testimony becomes a compact digest behind a privacy shield; only that digest reaches the cloud.}
\end{figure}

This convenience hides a cost. Sending raw video of a defendant, patient, or customer exposes their face and voice to a third party. Video is token-heavy, so the request is expensive. The verdict is opaque and hard to contest. In legal, clinical, and enterprise settings, "upload the suspect's video" is a non-starter.

This paper asks a narrow question. How much signal survives if we never send the video --- only a compact, interpretable digest extracted on-device? We study this on the \emph{Real-life Trial Deception} dataset \cite{ref1} and make four contributions.

\begin{itemize}
  \item \textbf{A privacy-preserving multimodal pipeline.} Whissle's on-device STT and vision stack turns each clip into \textasciitilde{}250 interpretable verbal, paralinguistic, and visual features, with no raw media leaving the box (see Method).
  \item \textbf{A clean four-way comparison} across two axes --- \emph{trained vs. zero-shot} and \emph{with-LLM vs. without-LLM} --- under speaker-independent evaluation (see Experiments). A trained classifier on our digest (AUC 0.741) matches an LLM watching raw video (0.749). An LLM judging our digest reaches 0.755, beating the video model.
  \item \textbf{A cost analysis} showing the digest uses 7.8$\times$ fewer LLM input tokens than raw video (see Cost Analysis). It is cheaper, faster, private, and more accurate.
  \item \textbf{An honesty audit.} We show the literature's 75\% is a leave-one-\emph{video}-out number inflated by speaker leakage. We quantify a demographic confound and report the speaker-independent result (see Ablation Studies).
\end{itemize}

\section{Related Work}

\textbf{The 2015 baseline.} Pérez-Rosas et al. introduced the \emph{Real-life Trial Deception} dataset and the first multimodal system on it \cite{ref1}. Earlier work used lab or crowdsourced lies --- mock crimes, the "werewolf" game \cite{ref4} --- where unmotivated subjects generalise poorly to real stakes. They instead used 121 public courtroom clips labelled from verdicts and exonerations \cite{ref1}. Their system fused transcript n-grams with 40 hand-coded MUMIN gesture features for 75.2\% accuracy --- but from a single best decision-tree cell; a random forest scored only 50.4\% \cite{ref1}.

\textbf{Verbal and non-verbal cues.} Two older strands feed this work. Text-only methods use n-grams and lexicons like LIWC: liars use fewer first-person and exclusive words and more negative emotion \cite{ref2}, each weak alone but useful together. Behavioural methods use physiological signals or hand-coded facial and gesture annotations \cite{ref3}; frowns, raised brows, and head turns track deception, and deceivers blink and shake their heads \emph{less} \cite{ref1,ref3}. These cues are noisy --- and in 2015 were labelled by hand with knowledge of each clip, not extracted automatically \cite{ref1}.

\textbf{The present: LLMs over video.} Multimodal LLMs now judge deception from video zero-shot \cite{ref6}, and new datasets pair audio with video \cite{ref5}. They are powerful but expensive and privacy-invasive --- and, as we show, no better than reasoning over a compact digest.

\textbf{The gap we address.} Three limits of the 2015 work persist. Its non-verbal features are hand-annotated, not automatic --- we extract them on-device. Its leave-one-\emph{video}-out evaluation leaks speaker identity and inflates accuracy --- we evaluate speaker-independently. And neither it nor video LLMs count token cost --- we measure it directly.

\section{Dataset}

The \emph{Real-life Trial Deception} dataset has \textbf{121 clips} (61 deceptive, 60 truthful), averaging \textasciitilde{}28 s, taken from public trials and labelled by verdict, acquittal, or exoneration.

\textbf{The speaker-leakage trap.} But it has only \textbf{33 speakers}: one defendant supplies \textbf{32 clips}, and \textbf{7 appear in both classes}. Under the original leave-one-\emph{video}-out split, holding out one such clip still leaves 31 of the same person in training --- so a model learns \emph{who} is speaking, not \emph{whether they lie}. We instead use \textbf{leave-one-speaker-out (LOSO)}, holding out all of a person's clips together. It scores lower, but it is the only protocol that measures generalisation to a new individual.

\begin{figure}[H]
\centering
\includegraphics[width=\linewidth]{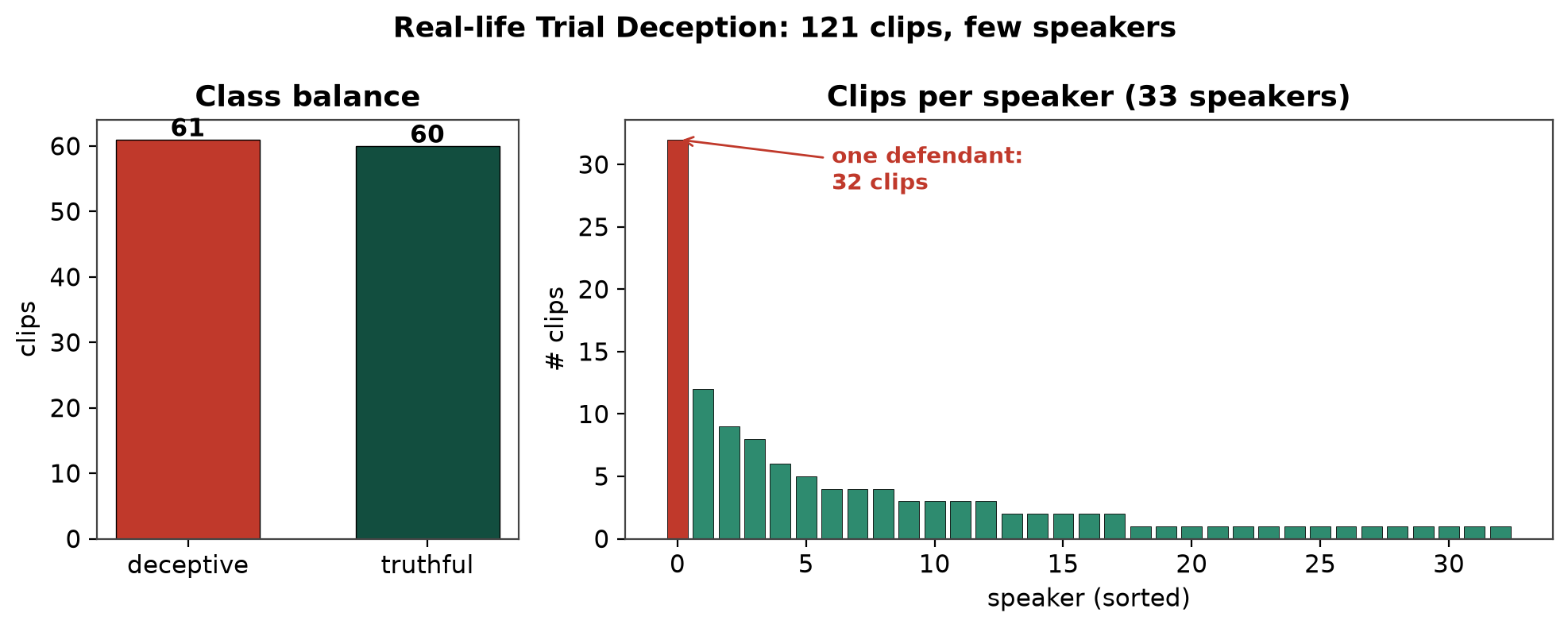}
\caption{Balanced 61/60 classes, but one defendant supplies 32 of 121 clips --- the source of speaker leakage.}
\end{figure}

\section{Method}

Three on-device lanes convert each clip into a feature digest. No raw media leaves the device.

\textbf{Text lane (Whissle STT).} The transcript, with probability distributions over \textbf{emotion, age, gender, and 33 intents}. A focused \textbf{deception-intent filter} scores labels such as denial, confession, justification, avoidance, and contradiction. A \textbf{speech-analysis} block covers fluency, grammar, vocabulary range, pitch, and rhythm (pause rate, speaking ratio, inter-word intervals). Psycholinguistic rates cover pronouns, negations, and hedges.

\textbf{Visual lane (audio-visual).} Per-frame facial emotion, gaze direction, head pose, blink, and hand gestures, aggregated into behavioural statistics: gaze aversion, head-motion / fidgeting, emotional variability, and blink rate.

\textbf{Acoustic lane (prosody).} Pitch (F0), jitter/shimmer, and pause statistics.

This yields \textasciitilde{}250 interpretable features per clip. We study four configurations along two axes (Table~\ref{tab:setup}):

\begin{table}[H]
\centering
\begin{tabular}{lll}
\toprule
 & No LLM & With LLM \\
\midrule
\textbf{Zero-shot} & majority baseline & LLM-as-judge over our digest (text-only); LLM over raw video \\
\textbf{Trained} & gradient-boosting on the digest & trained late-fusion of digest model + LLM \\
\bottomrule
\end{tabular}
\caption{The four system configurations, compared along two axes.}
\label{tab:setup}
\end{table}

\begin{figure}[H]
\centering
\includegraphics[width=\linewidth]{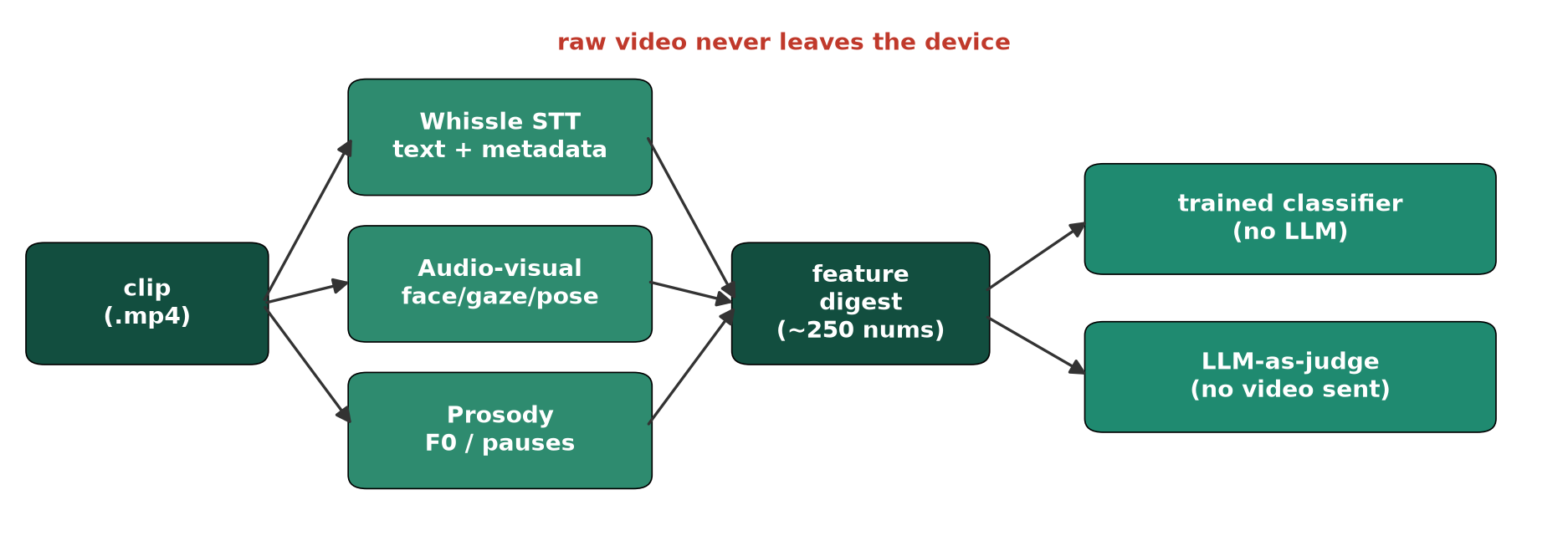}
\caption{Three local lanes build a \textasciitilde{}250-number digest; a classifier or LLM-as-judge predicts deception. Video never leaves the device.}
\end{figure}

\section{Experiments}

\textbf{Setup.} All numbers are leave-one-speaker-out, pooled out-of-fold over 121 clips. The trained model is gradient-boosted trees. The LLM judges are zero-shot, prompted neutrally with a base-rate anchor that weights verbal content over weak behavioural cues.

\textbf{Main results.} Table~\ref{tab:results} reports every system.

\begin{table}[H]
\centering
\begin{tabular}{llllll}
\toprule
System & Trained? & LLM? & Video sent? & Accuracy & ROC-AUC \\
\midrule
Majority baseline & --- & --- & No & 0.504 & 0.500 \\
Self-hosted features $\rightarrow$ gradient-boosting & Yes & No & No & 0.678 & 0.741 \\
LLM judges our features --- Claude Opus 4.8 & No & Yes & No & 0.620 & \textbf{0.755} \\
LLM judges our features --- Gemini 2.5 Pro & No & Yes & No & 0.669 & 0.704 \\
LLM watches raw video --- Gemini 2.5 Pro & No & Yes & Yes & 0.669 & 0.749 \\
Self-hosted + LLM (trained late fusion) & Yes & Yes & Yes & 0.678 & 0.752 \\
\bottomrule
\end{tabular}
\caption{Honest leave-one-speaker-out performance of every system.}
\label{tab:results}
\end{table}

Two independent roads reach \textasciitilde{}0.75 without exposing video. Train a small model on the digest (0.741), or hand the digest to a frontier LLM (0.704--0.755). The best single result --- \textbf{Claude Opus 4.8 over our digest, 0.755} --- exceeds the same model class watching raw video (0.749).

\textbf{Comparison to the literature.} Under the paper's own leave-one-video-out protocol, our features reproduce and exceed its numbers (0.752--0.777 vs. 0.752; Table~\ref{tab:protocol}). Under honest LOSO, everything drops by the size of the leakage.

\begin{table}[H]
\centering
\begin{tabular}{lll}
\toprule
Protocol & Original paper & Our features \\
\midrule
Leave-one-video-out (speaker-leaky) & 0.752 & 0.752--0.777 \\
Leave-one-speaker-out (honest) & not reported & 0.741 \\
\bottomrule
\end{tabular}
\caption{Our features under both cross-validation protocols.}
\label{tab:protocol}
\end{table}

\begin{figure}[H]
\centering
\includegraphics[width=\linewidth]{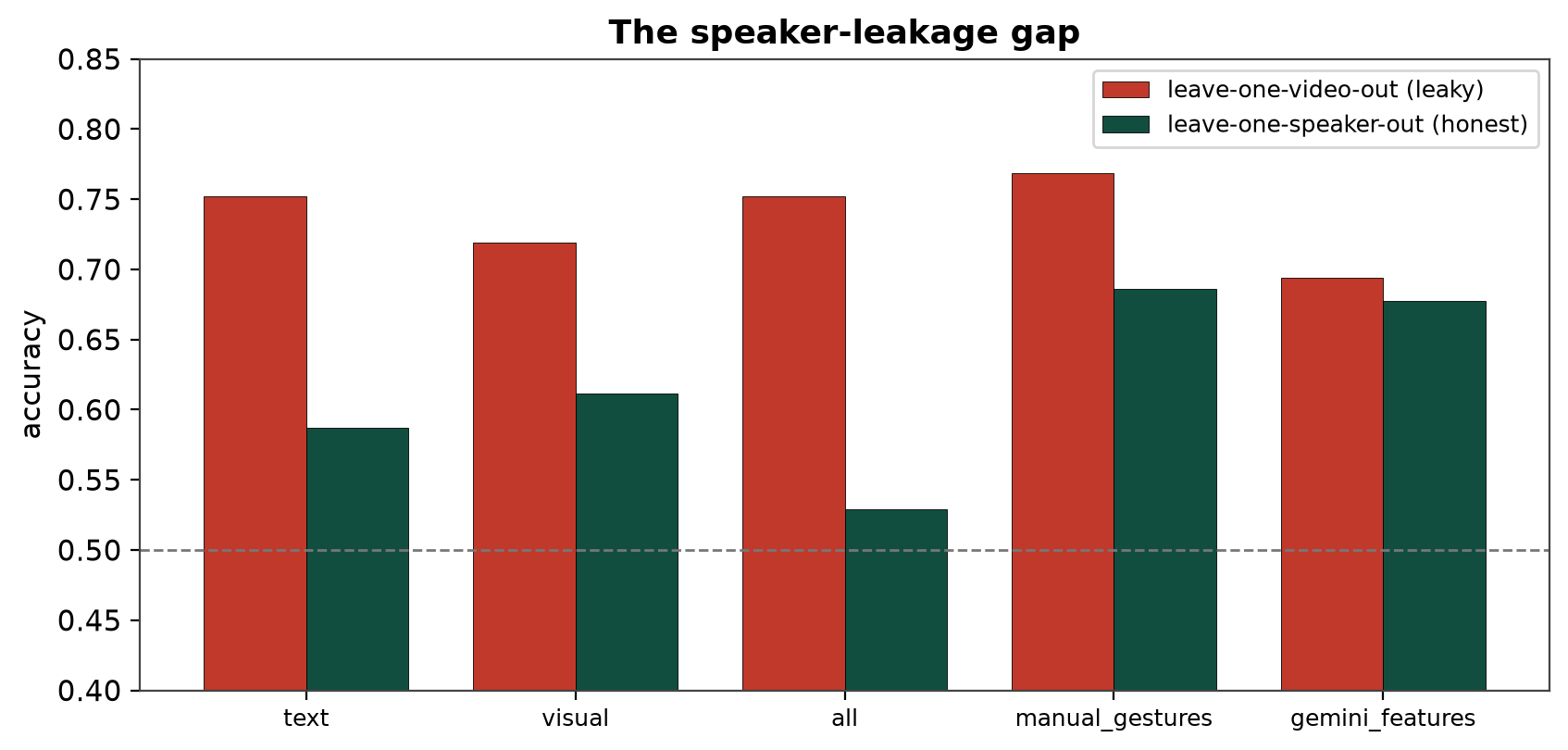}
\caption{Accuracy under leave-one-video-out (red) versus honest LOSO (green); the gap is the leakage inflation.}
\end{figure}

\begin{figure}[H]
\centering
\includegraphics[width=\linewidth]{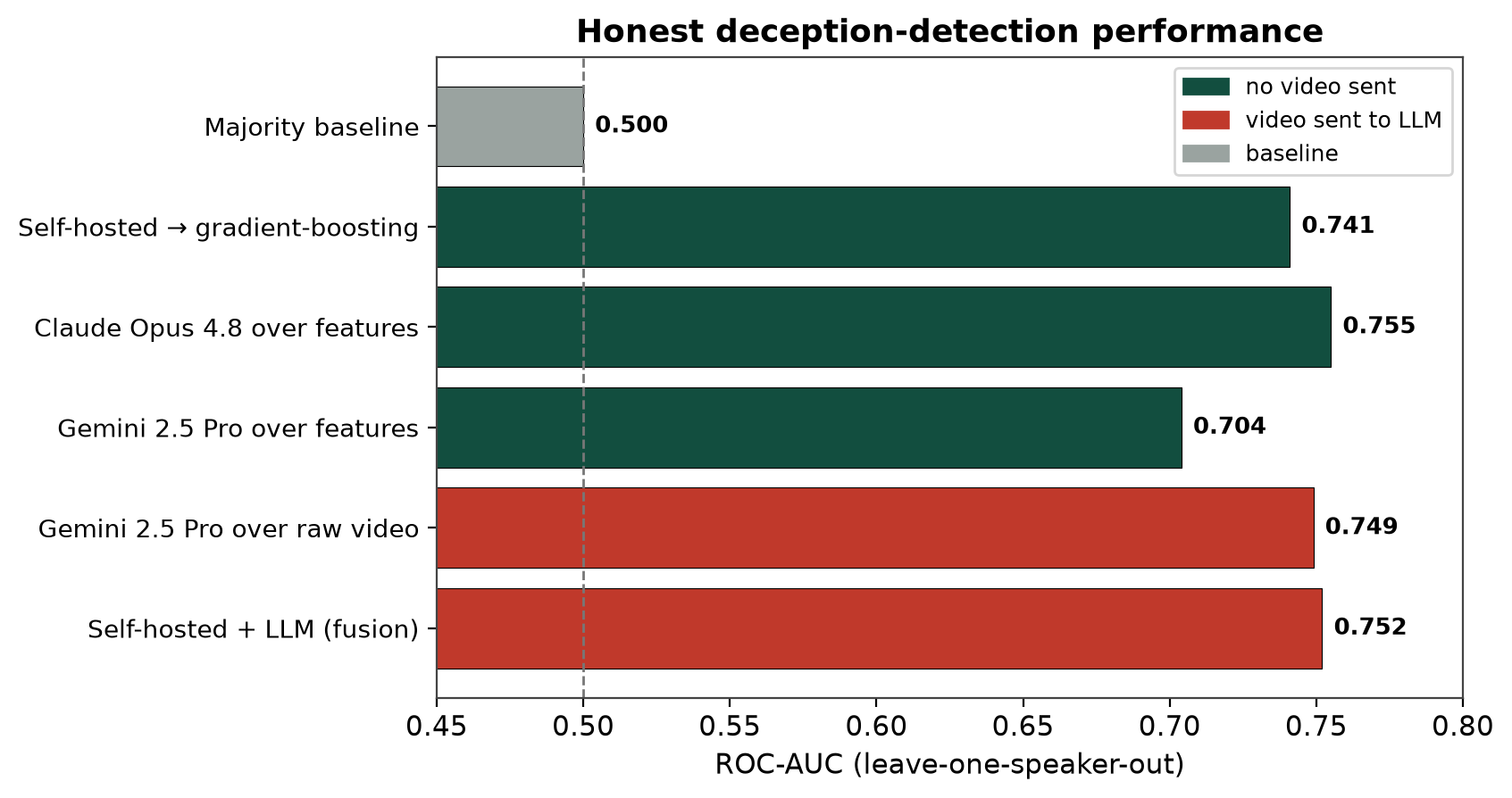}
\caption{Honest LOSO ROC-AUC for six systems. The best --- Claude Opus 4.8 over our digest (0.755) --- sends no video.}
\end{figure}

\section{Ablation Studies}

\textbf{Feature groups.} We partition the digest into seven groups and measure each alone, plus the AUC lost when it is removed (Table~\ref{tab:ablation}; LOSO, full set = 0.741):

\begin{table}[H]
\centering
\begin{tabular}{llll}
\toprule
Group & \# feats & Alone AUC & Marginal $\Delta$ (drop) \\
\midrule
STT metadata probs (emotion/age/gender/intent) & 119 & 0.700 & +0.039 \\
Visual (face/gaze/pose/gesture) & 42 & 0.579 & \textbf{+0.133} \\
Speech-analysis (fluency/grammar/rhythm) & 15 & 0.551 & +0.028 \\
Prosody (librosa) & 21 & 0.448 & +0.028 \\
Lexical (psycholinguistic) & 22 & 0.564 & +0.022 \\
Deception-intent filter & 15 & 0.518 & +0.023 \\
Speech structure (rate/pauses/conf) & 20 & 0.490 & $-$0.008 \\
\bottomrule
\end{tabular}
\caption{Each feature group alone, and the AUC lost when it is removed.}
\label{tab:ablation}
\end{table}

The STT-metadata group is strongest alone, but \textbf{the visual lane is the most complementary}. Removing it costs 0.133 AUC despite its modest solo score, because it adds signal no other lane captures.

\textbf{The visual fix.} Courtroom faces are small and oblique, and MediaPipe's default confidence rejected \textasciitilde{}half the frames. Lowering it raised face-detection from 0.50 to 0.80 and the visual lane from 0.61 to 0.67 AUC. This lifted the self-hosted system from 0.670 to 0.741.

\textbf{Demographic confound.} A few female defendants dominate the deceptive class, so the model's gender and age reads correlate with the label. Dropping all gender/age features lowers the self-hosted AUC to \textbf{0.678}, our bulletproof number. The with-LLM and direct-video results are unaffected by removing \emph{our} demographic features.

\textbf{Prompt sensitivity of the LLM judge.} A naive "forensic" prompt makes Gemini over-call deception (AUC 0.62, 79\% deceptive calls). Emphasising verbal content over weak behavioural cues, and anchoring the 50/50 base rate, recalibrates it to 0.704 at a 54\% deceptive rate. The two LLMs carry opposite priors. Gemini leans deceptive, Claude leans truthful. Claude's strong ranking (0.755) is therefore masked by a conservative threshold (raw accuracy 0.62, \textasciitilde{}0.70 thresholded).

\textbf{Which intents predict deception?} Whissle's STT emits a probability over 33 speech intents per clip, which we probe alongside the deception-intent filter. We correlate each intent's probability with the label (point-biserial \emph{r}, n = 121), and a clear pattern emerges (Figure 5). Clips lean \textbf{deceptive} when their intent is \textbf{response, deny, disagree, agree, request, or suggest} --- the reactive, denial-oriented register of someone fielding accusatory questions. They lean \textbf{truthful} when the intent is \textbf{reminisce or anecdote}, the genuine narrative recall of someone recounting events. We are deliberately cautious. Effect sizes are small (|r| $\le$ 0.28). Although 10 of 113 intents reach uncorrected p < 0.05 (vs. \textasciitilde{}6 expected by chance), \emph{none survive} multiple-comparison correction. No single intent is a standalone lie detector. What is meaningful is, first, the \textbf{coherent, theory-aligned direction} --- random noise would not place denial and deflection on the deceptive side and genuine recollection on the truthful side --- and second, that intents are the \textbf{single strongest feature family} in the ablation (Table~\ref{tab:ablation}). The signal lives in the \emph{distribution} of speech acts, not in any one; confirming individual intents needs a larger corpus.

\begin{figure}[H]
\centering
\includegraphics[width=\linewidth]{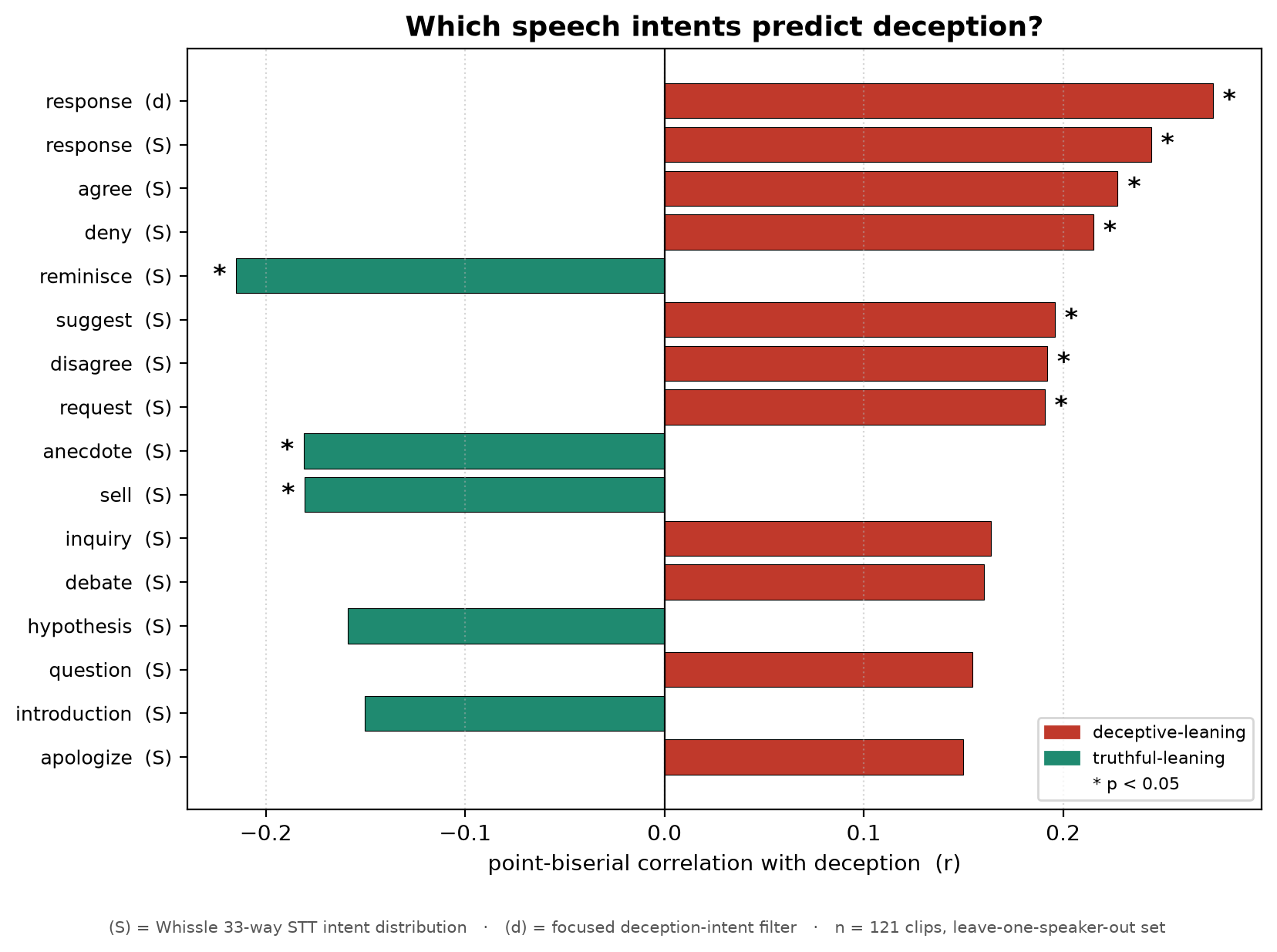}
\caption{Point-biserial correlation of each intent with the deceptive label; deny/disagree/response lean deceptive (red), reminisce/anecdote truthful (green). * marks p < 0.05.}
\end{figure}

\section{Cost Analysis}

Because we never send the video, the LLM processes far fewer tokens. We measure this with the provider's \texttt{count\_tokens} API over a sample of clips (Table~\ref{tab:cost}):

\begin{table}[H]
\centering
\begin{tabular}{lll}
\toprule
Input to the LLM & Mean input tokens & Relative \\
\midrule
Raw video ($\approx$296 tokens/s) & 9,810 & 1.0$\times$ \\
Our feature digest & 1,261 & \textbf{0.13$\times$ (7.8$\times$ fewer)} \\
\bottomrule
\end{tabular}
\caption{Mean input tokens per clip: raw video vs. our digest.}
\label{tab:cost}
\end{table}

At Gemini 2.5 Pro list pricing, this is \textasciitilde{}\$1.58 vs. \textasciitilde{}\$12.26 per 1,000 clips, an \textbf{\textasciitilde{}87\% input-cost reduction}. It also lowers latency, with no video decode or transfer, and, as shown, raises accuracy. Longer clips save more --- up to 10.7$\times$ on a 55 s clip.

\section{Limitations and Ethics}

This is a research probe on a tiny, US-trial-specific dataset, not a courtroom tool. No model here infers guilt; each predicts a dataset label derived from verdicts. Honest speaker-independent accuracy is \textasciitilde{}65--70\% --- far above chance, but far from proof. Part of the signal is a demographic confound that must be controlled. Deception detection must never be deployed to judge real people without rigorous, contestable, bias-audited validation. The interpretability of a feature-based system, where every cue is named, is a safeguard the opaque LLM verdict lacks.

\section{Conclusion}

You need not ship a person's video to a frontier model to detect deception. A compact, interpretable, on-device digest --- never exposing raw media --- matches a video-watching LLM with a small trained classifier (0.741). It \emph{exceeds} that model when the digest is handed to a frontier LLM judge (Claude Opus 4.8, 0.755), at 7.8$\times$ lower token cost. Privacy, cost, and accuracy point the same way: keep the video on the device.

\end{document}